\documentclass[a4paper, 10pt]{article}

\usepackage[a4paper,left=3cm,right=3cm,top=2.5cm,bottom=2.5cm]{geometry}
\usepackage{amsmath}
\usepackage{multirow}
\usepackage[table]{xcolor}
\usepackage{amsopn}
\usepackage{mathtools}
\usepackage{comment}
\usepackage{booktabs}
\usepackage{algorithm}
\usepackage{algpseudocode}
\usepackage{amssymb}
\usepackage{url}
\usepackage{bm}
\usepackage{tabularx}
\usepackage{enumerate}
\usepackage{enumitem}
\usepackage{array}
\usepackage{tikz}
\usepackage[export]{adjustbox}
\usepackage{tikzsymbols}
\usepackage{authblk}
\numberwithin{equation}{subsection}
\usepackage{xcolor}
\usepackage[hidelinks]{hyperref}
\usepackage{cleveref}
\usepackage{chngcntr}
\usepackage{xfrac}  
\usepackage{graphicx}
\usepackage{subcaption}
\usepackage{placeins}
\usepackage{float}

\begin{document}

\title{Trainable Spline Representations for Physics-Informed Learning}
\author[1]{Giovanni Canali\footnote{giovanni.canali@sissa.it}}
\author[1,2]{Nicola~Demo\footnote{nicola.demo@sissa.it}}
\author[1,2]{Gianluigi~Rozza\footnote{gianluigi.rozza@sissa.it}}

\affil[1]{Mathematics Area, mathLab, SISSA, via Bonomea 265, I-34136, Trieste, Italy}
\affil[2]{FAST Computing Srl, via Mazzini 20, I-34136, Trieste, Italy}
\maketitle

\begin{abstract}
   This work introduces Physics-Informed Splines (PI-Splines), a structured spline-based architecture for physics-informed learning. Instead of representing the solution of a differential equation with a neural network, PI-Splines directly parametrize the unknown field through a tensor-product B-spline expansion with trainable control coefficients. This formulation preserves the residual-based training paradigm of Physics-Informed Neural Networks while providing compact support, explicit smoothness control, analytical derivatives, and a direct geometric interpretation of the trainable parameters. When compatible with the spline representation, boundary conditions can be imposed strongly by fixing suitable boundary control coefficients.
The proposed method is evaluated on several benchmark problems of increasing difficulty and compared with standard physics-informed frameworks under matched governing equations, collocation sets, loss terms, and optimization procedures, so as to isolate the effect of the approximation architecture. Numerical experiments show that PI-Splines provide a competitive and stable alternative to neural physics-informed architectures, particularly in settings where structured representations, locality, and parameter efficiency are desirable.
\end{abstract}

\section{Introduction}\label{sec:intro}
Physics-Informed Neural Networks (PINNs)~\cite{RAISSI2019686, karniadakisphysics2021} have emerged as a widely studied framework for the numerical solution of differential equations. Their main idea is to approximate the unknown solution by a neural network and to train its parameters by minimizing a loss function that penalizes the residual of the governing equation, together with initial, boundary, and, when available, data-misfit terms. This formulation provides a flexible mesh-free paradigm for forward and inverse problems, and has attracted considerable interest in computational science and engineering due to its ability to combine physical constraints with sparse or noisy observations.

Despite this flexibility, the use of PINNs as general-purpose solvers remains challenging. Their performance is often sensitive to the choice of network architecture, initialization, collocation points, loss weights, and optimizer~\cite{doi:10.1137/20M1318043, Wang2023AnEG}. In addition, standard multilayer perceptrons may exhibit spectral bias, which can hinder the representation of high-frequency components, sharp gradients, boundary layers, and multiscale structures~\cite{Rahaman2018OnTS}. Further difficulties arise from the simultaneous optimization of multiple loss terms, whose magnitudes and convergence rates may be poorly balanced~\cite{WANG2022110768}. As a consequence, although PINNs are attractive for data assimilation and inverse modeling, their accuracy and computational efficiency are not yet consistently competitive with established discretization-based methods such as the Finite Element Method (FEM) or the Finite Volume Method (FVM) for forward problems requiring high precision~\cite{pinns_vs_FEM}.

Several recent developments have sought to improve the expressiveness, robustness, and interpretability of neural solvers for differential equations. These include adaptive sampling strategies~\cite{daw2023mitigating}, modified loss formulations~\cite{MCCLENNY2023111722, ANAGNOSTOPOULOS2024116805}, domain-decomposition approaches~\cite{jagtap2020extended}, and alternative neural architectures~\cite{Wang2024PirateNetsPD}. Fourier-feature PINNs~\cite{WANG2021113938} address the spectral bias of standard multilayer perceptrons by enriching the input representation with oscillatory features before applying the neural network. This can improve the approximation of high-frequency and multiscale solutions, but the trial space remains a global neural parametrization. Consequently, the number of trainable parameters, the dependence on automatic differentiation, and the weak enforcement of boundary conditions through loss penalties are not fundamentally altered.

Among alternative architectures, Kolmogorov--Arnold Networks (KANs)~\cite{Toscano2024FromPT, liu2025kan} have recently received attention as a spline-based alternative to multilayer perceptrons, replacing fixed activation functions with learnable univariate non-linearities parametrized by splines on the edges of the network. This construction introduces an explicit functional representation inside the architecture and has shown promising behavior in scientific machine learning tasks. Nevertheless, KANs remain neural-network models, with nontrivial architectural choices, training dynamics, and computational costs associated with the evaluation and optimization of many spline-parametrized components. Moreover, the spline functions are internal components of the network map, rather than a direct approximation space for the solution field.

Spline-based parametrizations have also recently been explored in physics-informed learning. Wandel et al. introduced Spline-PINN, which combines Hermite-spline interpolation with convolutional representations for data-free PDE solution on structured grids~\cite{hermite}. Wang et al. proposed physics-informed deep B-spline networks, where a neural network predicts B-spline control points for parametric PDE solution approximation~\cite{wang2026physicsinformed}. Saidaoui et al. developed Deep NURBS, using NURBS-based admissible constructions to encode geometry and boundary information in physics-informed neural networks~\cite{deepnurbs}. Sun et al. introduced PiSL, where spline learning is coupled with sparse regression for the data-driven identification of nonlinear dynamical systems~\cite{SUN2023110165}. In a related geometry-oriented direction, Tamburlin et al. proposed a differentiable CAD-native framework in which a neural displacement field acts directly on NURBS control points and physically meaningful integral constraints are evaluated through analytical B-spline derivatives and numerical quadrature \cite{tamburlin2026constraint}. These approaches demonstrate that spline and NURBS representations can improve smoothness, differentiability, boundary treatment, geometric consistency, and data efficiency. However, they either retain a neural network as the primary trainable map, use splines as interpolation or reconstruction devices, focus on geometry parametrization or admissible neural trial functions, or address system identification rather than directly optimizing a spline trial space for forward physics-informed solution approximation.

In this work, we investigate a different use of splines for physics-informed learning. Rather than embedding splines inside a neural architecture, we directly parametrize the trial solution by a tensor-product B-spline expansion~\cite{splinebook, nurbsbook} and optimize its control coefficients through a physics-informed objective. The resulting method, referred to as Physics-Informed Splines (PI-Splines), preserves the residual-based training principle of PINNs while replacing the neural-network ansatz with a structured spline approximation. Boundary conditions can be imposed strongly by fixing suitable boundary control coefficients, thereby removing the corresponding boundary penalty from the loss whenever the prescribed data and the spline representation are compatible.

This formulation provides several advantages over standard neural physics-informed parametrizations. First, the number of trainable parameters is explicitly controlled by the number of unfrozen control coefficients and can be substantially smaller than that of multilayer perceptrons, Fourier-feature PINNs, or spline-augmented neural architectures. Second, B-spline basis functions have compact support, so each control coefficient affects the solution only locally. This yields a direct geometric correspondence between regions of the control grid and regions of the solution field, improving interpretability and enabling localized corrections of the approximation. Third, the derivatives required by the governing equations are available analytically from the spline basis. Consequently, the physics residual can be evaluated through explicit spline derivative formulas, without numerical differentiation and without propagating differential operators through a deep neural network by automatic differentiation. Finally, the smoothness of the approximation is controlled directly through the spline order and knot multiplicities, rather than being an implicit consequence of network depth, activation choice, or training dynamics.

The proposed approach should not be interpreted as a universal approximator at a fixed discretization. For a prescribed knot vector, spline order, and number of control coefficients, the approximation space is finite-dimensional and therefore cannot represent arbitrary functions. Universality holds only at the level of a sequence of refined spline spaces: as the number of control coefficients increases and the mesh size tends to zero, spline spaces provide dense approximation families in standard function spaces. The aim of this work is therefore not to maximize expressiveness through overparametrization, but to assess whether a compact, structured, and physics-informed spline space can provide a favorable trade-off between accuracy, robustness, interpretability, and computational cost for representative differential problems.

The remainder of the paper is organized as follows. Section~\ref{sec:methods} introduces the methodological framework, reviewing the PINN formulation, the tensor-product B-spline representation, analytical spline derivatives, and the construction of PI-Splines with learnable control coefficients. Section~\ref{sec:results} describes the benchmark problems, presents the numerical results and compares PI-Splines with standard PINN baselines. In particular, Section~\ref{subsec:sensitivity_analysis} analyzes the sensitivity of the method to the spline order and the number of control coefficients. Finally, Section~\ref{sec:conclusions} summarizes the main findings, discusses computational trade-offs, and outlines directions for future work.

\section{Methodology}\label{sec:methods}
This section introduces the proposed methodology. We first review the Physics-Informed Neural Network (PINN) formulation and the B-spline representation, focusing on the aspects that are subsequently used to define the spline-based framework: residual-based training, tensor-product spline approximation, analytical differentiation, and boundary control through endpoint coefficients. We then combine these elements into Physics-Informed Splines (PI-Splines), where the neural-network ansatz is replaced by a trainable spline expansion whose control coefficients define the optimization variables. The resulting approach preserves the physics-informed structure of PINNs while exploiting the locality, smoothness, and geometric interpretability of the B-spline parametrization.

\subsection{Physics-Informed Neural Networks}
\label{subsec:pinns}

PINNs approximate solutions to differential equations by embedding the governing physical laws directly into the training process. Let $\Omega \subset \mathbb{R}^d$ be the spatial domain and let $T>0$. We consider a time-dependent problem of the form
\begin{equation}
    \frac{\partial}{\partial t} u(\mathbf{x},t) - \mathcal{N}[u](\mathbf{x},t) = 0,
    \qquad(\mathbf{x},t)\in \Omega \times (0,T],
    \label{eq:pde_time_dependent}
\end{equation}
where $u:\Omega\times[0,T]\to\mathbb{R}^q$ is the unknown solution and $\mathcal{N}$ is a possibly nonlinear differential operator involving spatial derivatives. The problem is completed with an initial condition and boundary conditions:
\begin{equation}
    u(\mathbf{x},0)=u_0(\mathbf{x})\quad \mathbf{x}\in\Omega,
    \qquad\quad \mathcal{B}[u](\mathbf{x},t)=g(\mathbf{x},t)
    \quad (\mathbf{x},t)\in\partial\Omega\times(0,T],
    \label{eq:pinn_ic_bc}
\end{equation}
where $\mathcal{B}$ denotes the boundary operator.

In the PINN framework, the solution $u$ is approximated by a neural network $u_\theta(\mathbf{x},t)$, with trainable parameters $\theta$. Substituting $u_\theta$ into the governing equation defines the physics residual
\begin{equation}
    r_\theta(\mathbf{x},t) = \frac{\partial}{\partial t} u_{\theta}(\mathbf{x},t)
    - \mathcal{N}[u_\theta](\mathbf{x},t).
    \label{eq:pinn_residual}
\end{equation}
The parameters are optimized by minimizing a loss function that penalizes the residual, the violation of the initial and boundary conditions, and, when available, the mismatch with observational or high-fidelity data:
\begin{equation}
    \mathcal{L}(\theta) =
    \lambda_r \mathcal{L}_r(\theta) + \lambda_i \mathcal{L}_i(\theta)
    + \lambda_b \mathcal{L}_b(\theta) + \lambda_d \mathcal{L}_d(\theta).
    \label{eq:pinn_total_loss}
\end{equation}
Here, $\lambda_r,\lambda_i,\lambda_b,\lambda_d\geq 0$ weight the relative importance of the different contributions. Given interior collocation points $\mathcal{X}_r = \{(\mathbf{x}_r^{(j)},t_r^{(j)})\}_{j=1}^{N_r} \subset \Omega\times(0,T]$, the residual loss is typically approximated by
\begin{equation}
    \mathcal{L}_r(\theta) = \frac{1}{N_r} \sum_{j=1}^{N_r}
    \left\lVert r_\theta(\mathbf{x}_r^{(j)},t_r^{(j)}) \right\rVert^2.
    \label{eq:pinn_residual_loss}
\end{equation}
The initial and boundary losses are defined analogously by evaluating the mismatch with the prescribed initial and boundary data. If measurements or reference solutions are available at points $\mathcal{X}_d = \{(\mathbf{x}_d^{(j)},t_d^{(j)})\}_{j=1}^{N_d}$, an additional supervised term can be included:
\begin{equation}
    \mathcal{L}_d(\theta) = \frac{1}{N_d} \sum_{j=1}^{N_d}
    \left\lVert u_\theta(\mathbf{x}_d^{(j)},t_d^{(j)}) - u^{(j)} \right\rVert^2.
    \label{eq:pinn_data_loss}
\end{equation}
The derivatives appearing in the residual are computed by automatic differentiation, which allows the differential operator to be evaluated directly within the computational graph. This makes PINNs mesh-free, since the residual can be imposed at arbitrary points in the space-time domain.

Although PINNs provide a flexible framework for combining physical constraints and data, their training remains challenging. Standard multilayer perceptrons may suffer from spectral bias, making high-frequency components, sharp gradients, boundary layers, and multiscale features difficult to approximate. Moreover, the different loss terms in \eqref{eq:pinn_total_loss} may have incompatible magnitudes or optimization dynamics, so reducing one term does not necessarily improve the others. As a result, the performance of PINNs can be sensitive to the choice of loss weights, collocation points, network architecture, initialization, and optimizer.

\subsection{B-spline representation}
\label{subsec:bspline_representation}

B-splines provide a flexible and compact representation of smooth functions through piecewise-polynomial basis functions. In the univariate case, let $\mathcal{X} = \{x_1,x_2,\ldots,x_m\}$ be a non-decreasing knot vector and let $k$ denote the spline order. The corresponding B-spline basis functions $B_{i,k}$ are piecewise polynomials of degree $p=k-1$, with compact support on the interval $[x_i,x_{i+k}]$. If $n$ denotes the number of control coefficients, then the length of the knot vector satisfies $m = n + k$. Thus, once the spline order and the number of control coefficients are prescribed, the number of knots is determined accordingly.

A univariate B-spline function is defined as
\begin{equation}
    S(x) = \sum_{i=1}^{n} C_i B_{i,k}(x), \qquad x \in [x_1,x_m],
    \label{eq:univariate_bspline}
\end{equation}
where $C_i\in\mathbb{R}$ are the control coefficients. These coefficients determine the shape of the spline but are not, in general, interpolated by the function. Interpolation at selected points may occur only under specific knot multiplicities. In particular, if the first and last knots are repeated $k$ times, the spline is clamped and interpolates the first and last control coefficients:
\begin{equation}
    S(x_1)=C_1, \qquad S(x_m)=C_n.
\end{equation}
The spline is considered on the parametric interval defined by the nonzero support of the basis functions, and the basis functions vanish outside their supports.

The B-spline basis functions are constructed recursively using the Cox--de Boor formula. For order $k=1$, the basis functions are piecewise constant:
\begin{equation}
    B_{i,1}(x) =
    \begin{cases}
        1, & x_i \leq x < x_{i+1}, \\
        0, & \text{otherwise}. \\
    \end{cases}
    \label{eq:bspline_order_one}
\end{equation}
For $k>1$, they are defined recursively as
\begin{equation}
    B_{i,k}(x) = \frac{x-x_i}{x_{i+k-1}-x_i} B_{i,k-1}(x)
    + \frac{x_{i+k}-x}{x_{i+k}-x_{i+1}} B_{i+1,k-1}(x),
    \label{eq:cox_de_boor}
\end{equation}
with the convention that terms with zero denominator are set to zero. This recursive construction yields basis functions that are non-negative, locally supported, and form a partition of unity on the spline domain, under the usual assumptions on the knot vector. The locality of the basis is one of the defining properties of B-splines: each coefficient $C_i$ affects the spline only on the support of $B_{i,k}$, namely $[x_i,x_{i+k}]$. This gives the representation a sparse and geometrically interpretable structure.

The smoothness of a B-spline depends on both the degree and the knot multiplicities. At a knot with multiplicity $r$, a spline of degree $p$ is generally $C^{p-r}$-continuous. Therefore, repeated knots reduce regularity and can be used to introduce lower smoothness or interpolation properties at selected locations. Conversely, simple interior knots yield the highest continuity allowed by the polynomial degree, namely $C^{p-1}$.

\subsubsection{Tensor-product extension}
\label{subsubsec:tensor_product_bspline}

The univariate B-spline construction extends naturally to higher-dimensional domains by means of tensor products. In the bivariate case, let $\mathcal{X} = \{x_1,x_2,\ldots,x_m\}, \mathcal{Y} = \{y_1,y_2,\ldots,y_l\}$ be two non-decreasing knot vectors, and let $k$ and $s$ denote the spline orders in the $x$- and $y$-directions, respectively. A bivariate B-spline surface is defined as
\begin{equation}
    S(x,y) = \sum_{i=1}^{n_x} \sum_{j=1}^{n_y} C_{ij} B_{i,k}(x) B_{j,s}(y),
    \qquad (x,y)\in [x_1,x_m]\times[y_1,y_l],
    \label{eq:bivariate_bspline}
\end{equation}
where $C\in\mathbb{R}^{n_x\times \,n_y}$ is the matrix of control coefficients. The entries $C_{ij}$ influence the shape of the surface but are not generally interpolated, except under suitable knot multiplicities.

The tensor-product structure separates the construction along each coordinate direction. The functions $B_{i,k}(x)$ and $B_{j,s}(y)$ are standard univariate B-spline basis functions associated with the knot vectors $\mathcal{X}$ and $\mathcal{Y}$. This makes the multidimensional construction modular: each direction can have its own degree, knot spacing, and boundary multiplicities. The same approach extends directly to higher dimensions by introducing one knot vector and one univariate B-spline basis per coordinate direction. For a $d$-dimensional domain with coordinates $\mathbf{x}=(x_1,\ldots,x_d)$, a tensor-product B-spline can be written as
\begin{equation}
    S(\mathbf{x}) = \sum_{i_1=1}^{n_1} \cdots \sum_{i_d=1}^{n_d}
    C_{i_1,\ldots,i_d} \prod_{\alpha=1}^{d} B_{i_\alpha,k_\alpha}^{(\alpha)}(x_\alpha),
    \label{eq:multivariate_bspline}
\end{equation}
where $C$ is a tensor of control coefficients and $B_{i_\alpha,k_\alpha}^{(\alpha)}$ denotes the univariate B-spline basis in the $\alpha$-th coordinate direction. Thus, the multidimensional construction is obtained by separable products of univariate bases.

\subsubsection{Analytical derivatives}
\label{subsubsec:bspline_derivatives}

Since B-splines are piecewise-polynomial functions, their derivatives can be computed analytically. The derivative of a univariate B-spline basis function of order $k$ is
\begin{equation}
    \frac{d}{dx} B_{i,k}(x) = \frac{k-1}{x_{i+k-1}-x_i} B_{i,k-1}(x)
    -\frac{k-1}{x_{i+k}-x_{i+1}} B_{i+1,k-1}(x),
    \label{eq:bspline_basis_derivative}
\end{equation}
where terms with zero denominator are again set to zero. Therefore, the derivative of the spline is obtained by differentiating the basis functions:
\begin{equation}
    S'(x)=\sum_{i=1}^{n}C_i\frac{d}{dx}B_{i,k}(x).
\end{equation}
Higher-order derivatives follow by repeated application of \eqref{eq:bspline_basis_derivative}, up to the polynomial degree of the spline. Equivalently, the derivative of a B-spline can itself be written as a lower-degree B-spline expansion. For degree $p=k-1$,
\begin{equation}
    S'(x)=\sum_{i=1}^{n-1} D_i B_{i,k-1}(x),
\end{equation}
where the derivative control coefficients are
\begin{equation}
    D_i = p\ \frac{C_{i+1} - C_i}{x_{i+k}-x_{i+1}}.
    \label{eq:derivative_control_points}
\end{equation}
This expression highlights that differentiation maps a spline of degree $p$ to a spline of degree $p-1$, with new coefficients given by scaled finite differences of the original control coefficients.

For tensor-product splines, partial derivatives are obtained by differentiating only the basis associated with the corresponding coordinate direction. Higher-order and mixed derivatives are constructed analogously.

This analytical differentiability is a key feature of spline representations. Differential operators can be evaluated explicitly in terms of polynomial basis functions and their derivatives, without changing the structure of the approximation space. At the same time, the compact support of the basis functions ensures that only a small number of terms contribute at each evaluation point, yielding sparse and local computations.

\subsubsection{Boundary control through endpoint coefficients}
\label{subsubsec:boundary_control_bspline}

Clamped B-splines provide direct control over endpoint values. In the univariate case, when the first and last knots are repeated $k$ times, the spline interpolates the first and last control coefficients:
\begin{equation}
    S(x_1)=C_1, \qquad S(x_m)=C_n.
\end{equation}
This property allows Dirichlet boundary values to be prescribed directly by fixing the corresponding endpoint coefficients. Leveraging \eqref{eq:derivative_control_points}, the endpoint derivatives can be expressed directly in terms of the boundary control coefficients. In particular, for a clamped univariate B-spline, the first derivative at the left and right endpoints is given by 
\begin{equation}
    S'(x_1) = p\ \frac{C_2 - C_1}{x_{k+1}-x_1},
    \qquad S'(x_m) = p\ \frac{C_n - C_{n-1}}{x_m-x_{m-k}},
    \label{eq:endpoint_derivatives}
\end{equation}
assuming the corresponding knot intervals are nonzero. Therefore, Neumann boundary values can be imposed by controlling differences between adjacent endpoint coefficients.

The same principle applies dimension-wise to tensor-product splines. In a bivariate spline, a Dirichlet condition on the boundary $x=x_1$ is controlled by the first row of coefficients $C_{1j}$, while a Neumann condition in the $x$-direction depends on the differences $C_{2j}-C_{1j}$. Similarly, boundary conditions on $x=x_m$, $y=y_1$, and $y=y_l$ are controlled by the corresponding boundary rows or columns of the control coefficient matrix. Thus, both Dirichlet and Neumann information can be encoded directly through boundary layers of the control grid.

\subsection{Physics-Informed Splines}
\label{subsec:pi_splines}

PI-Splines replace the neural-network ansatz of PINNs with a tensor-product B-spline representation, while preserving the physics-informed training principle introduced in Section~\ref{subsec:pinns}. The key difference is therefore not in the structure of the residual-based objective, but in the parametrization of the trial solution. Instead of learning the weights and biases of a neural network, the proposed method learns the control coefficients of a spline expansion.

Let $\Omega\subset\mathbb{R}^d$ denote the spatial domain. The approximate solution is represented as
\begin{equation}
    u_C(\mathbf{x}) = \sum_{i_1=1}^{n_1} \cdots \sum_{i_d=1}^{n_d}
    C_{i_1,\ldots,i_d} \prod_{\alpha=1}^{d} B_{i_\alpha,k_\alpha}^{(\alpha)}(x_\alpha),
\end{equation}
where $C$ is the tensor of control coefficients and $B_{i_\alpha,k_\alpha}^{(\alpha)}$ is the univariate B-spline basis associated with the $\alpha$-th coordinate direction. In time-dependent problems, time is treated as an additional coordinate, so that the same tensor-product construction is applied over the space-time domain. The trainable parameters of the model are the entries of $C$ that are not fixed by prescribed constraints.

This formulation turns the control grid into the optimization variable of the physics-informed problem. The residual, initial-condition, boundary-condition, and data terms are evaluated as in the PINN formulation, with the neural approximation $u_\theta$ replaced by the spline approximation $u_C$. The only structural modification concerns boundary conditions that can be imposed directly through the spline representation, as described in \ref{subsubsec:boundary_control_bspline}, by fixing suitable boundary layers of the control tensor. In that case, the corresponding boundary control coefficients are fixed a priori, excluded from the set of learnable parameters, and the boundary penalty is no longer required. The optimization objective therefore becomes
\begin{equation}
    \mathcal{L}(C) = \lambda_r \mathcal{L}_r(C) + 
    \lambda_i \mathcal{L}_i(C) + \lambda_d \mathcal{L}_d(C),
\end{equation}
where the optimization is performed only over the unfrozen coefficients.

The exact imposition of boundary conditions removes the need to balance the residual loss against a boundary penalty and ensures that the admissible functions satisfy the prescribed boundary data by construction. Initial conditions can, in principle, be imposed in the same way by fixing the control coefficients associated with the initial time boundary. However, in the time-dependent experiments considered in this work, this hard enforcement strategy led to less stable optimization. We therefore impose spatial boundary conditions strongly whenever possible, while retaining a weak penalty formulation for the initial condition.

A further distinguishing feature of PI-Splines is the explicit availability of the derivatives required by the governing equation. Since B-splines are piecewise-polynomial functions, spatial and temporal derivatives can be computed by differentiating the basis functions or, equivalently, by applying the corresponding derivative-control-coefficient formulas. The physics residual is therefore evaluated directly from the analytical derivatives of the spline expansion, without numerical differentiation and without relying on automatic differentiation through a deep neural network. As a result, the evaluation of the residual does not introduce discretization errors associated with derivative approximation.

In addition, the compact support of B-spline basis functions endows the parametrization with a local and interpretable structure. Each control coefficient influences the solution only on the support of the associated tensor-product basis function and therefore affects a restricted region of the domain. The learned parameters consequently retain a direct geometric meaning: localized subsets of the control grid correspond to localized regions of the solution field. This property can be advantageous for representing spatially confined structures, sharp gradients, or localized solution features, which may be difficult to capture with globally parametrized neural-network ansatzes. In standard PINNs, individual weights typically contribute to the approximation throughout the domain, making the relation between parameters and local solution behavior less explicit.

The approximation properties of PI-Splines are governed by the knot vectors and spline orders in each coordinate direction. These choices determine the number of control coefficients, the local polynomial degree, the continuity across knot spans, and the computational cost of evaluating the approximation and its derivatives. Anisotropic choices of knots or spline orders may be used to allocate resolution preferentially along selected spatial or temporal directions. In the numerical experiments presented here, however, we adopt uniform knot distributions and spline orders unless otherwise specified. A sensitivity analysis with respect to the number of knots and the spline order is reported in Section~\ref{subsec:sensitivity_analysis}.

Compared with standard PINNs, PI-Splines can require substantially fewer trainable parameters, since only the unfrozen entries of the control tensor are optimized. In the numerical experiments reported in this work, this reduction can reach up to two orders of magnitude relative to classical multilayer perceptrons. This compact parametrization reduces the dimension of the optimization problem and preserves the geometric interpretability of the control grid, while retaining a smooth approximation space.

\section{Results}\label{sec:results}
We evaluate the proposed architecture on a set of differential problems spanning smooth low-frequency solutions, anisotropic oscillations, localized sharp gradients, and time-dependent wave propagation. The objective is to determine whether replacing the neural-network ansatz by a tensor-product B-spline expansion improves accuracy, parameter efficiency, and robustness while preserving the standard physics-informed training paradigm.

All experiments are implemented in PINA~\cite{coscia2023physics} and run on a single NVIDIA A$800$ $40$GB Active GPU. The implementation of PI-Splines and the scripts used to reproduce the numerical experiments are publicly available at \url{https://github.com/GiovanniCanali/PI-Splines.git}. PI-Splines are compared with three physics-informed baselines: standard PINNs, PINNs with Fourier-feature embeddings, and physics-informed Kolmogorov--Arnold Networks, denoted here as PIKANs. Across methods, the governing equation, collocation sets, loss terms, optimization schedule, and training strategy are kept fixed whenever possible, so that the comparison primarily reflects the choice of approximation architecture. Analytical solutions are available for all benchmarks and are used to compute the mean absolute error on a fixed evaluation grid, independent of the training collocation set. The same optimization procedure is used for all models and consists of an Adam~\cite{2015-kingma} phase followed by LBFGS refinement, with problem-specific settings reported in the benchmark descriptions and in Appendix~\ref{sec:app_A}. The remainder of the section is organized as follows. Section~\ref{subsec:benchmark_problems} introduces the benchmark problems, Section~\ref{subsec:quantitative_results} presents the quantitative comparison, and Section~\ref{subsec:sensitivity_analysis} examines the sensitivity of PI-Splines to spline order and number of control points.

\subsection{Benchmark problems}
\label{subsec:benchmark_problems}

The benchmark suite probes regimes of increasing numerical difficulty. The Poisson problem provides a smooth low-frequency reference case. The Helmholtz problem introduces direction-dependent frequency content, requiring the approximation space to resolve different spatial frequencies along different coordinate directions. The exponential problem contains localized sharp gradients induced by rapidly varying exponential factors, while the acoustic wave problem extends the comparison to a time-dependent setting with competing residual, initial-condition, and boundary-condition constraints.

For PI-Splines, homogeneous Dirichlet boundary conditions are imposed strongly by fixing the corresponding boundary control coefficients of the clamped tensor-product spline basis. These coefficients are excluded from the trainable parameter set. The spline order and number of control points are selected separately for each benchmark and reported in the corresponding subsections.

\subsubsection{Poisson problem}
\label{subsubsec:poisson_problem}

First, we consider a two-dimensional Poisson problem defined over the unit square $\Omega=[0,1]\times[0,1]$ subject to homogeneous Dirichlet boundary conditions:
\begin{equation}
    \begin{aligned}
        \Delta u(x, y) = 2\pi^2 \sin(\pi x)\sin(\pi y) & \qquad \forall (x, y) \in \Omega, \\
        u(x, y) = 0 & \qquad \forall (x, y) \in \partial \Omega.
    \end{aligned}
\end{equation}
The analytical solution is $u(x,y) = -\sin(\pi x)\sin(\pi y)$ and is used as the reference field for error evaluation.

\begin{figure}[ht]
\centering
    \includegraphics[width=0.95\textwidth]{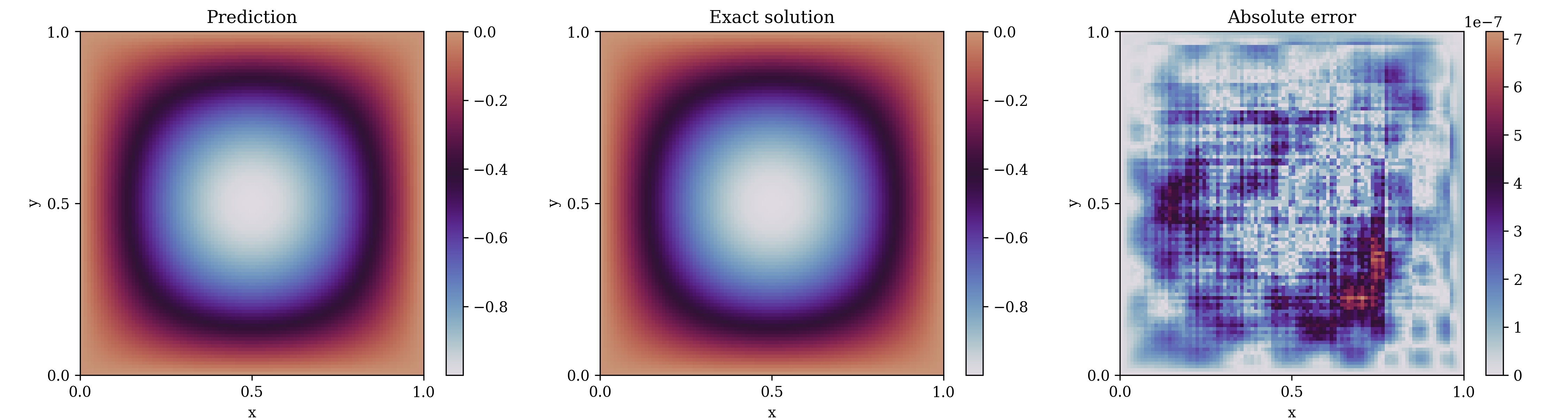}
    \caption{Poisson problem. \textbf{Left}: PI-Spline prediction. \textbf{Middle}: reference solution. \textbf{Right}: pointwise absolute error.}
    \label{fig:poisson_task}
\end{figure}

This problem provides a smooth, low-frequency test case for evaluating the accuracy of the spline parametrization in a regular elliptic setting. The PI-Spline approximation uses splines of order $6$ with $15$ control points in each coordinate direction and is trained for $5000$ epochs using $5000$ collocation points. Figure~\ref{fig:poisson_task} reports the reference solution, the PI-Spline prediction, and the corresponding pointwise absolute error.

\subsubsection{Helmholtz problem}
\label{subsubsec:helmholtz_problem}

The second benchmark is a Helmholtz problem with anisotropic frequency content, posed on $\Omega=[-1,1]\times[-1,1]$, following~\cite{SI2026114713}:
\begin{equation}
    \begin{aligned}
        \Delta u(x, y) + k^2 u(x, y) = f(x, y) & \qquad \forall (x, y) \in \Omega, \\
        u(x, y) = 0 & \qquad \forall (x, y) \in \partial \Omega,
    \end{aligned}
\end{equation}
where $f(x,y)=\left(k^2-a_1^2\pi^2-a_2^2\pi^2\right)\sin(a_1\pi x)\sin(a_2\pi y)$. We set $k=1$, $a_1=1$, and $a_2=4$, yielding the analytical solution
$u(x,y)=\sin(\pi x)\sin(4\pi y)$.

\begin{figure}[ht]
    \centering
    \includegraphics[width=0.95\textwidth]{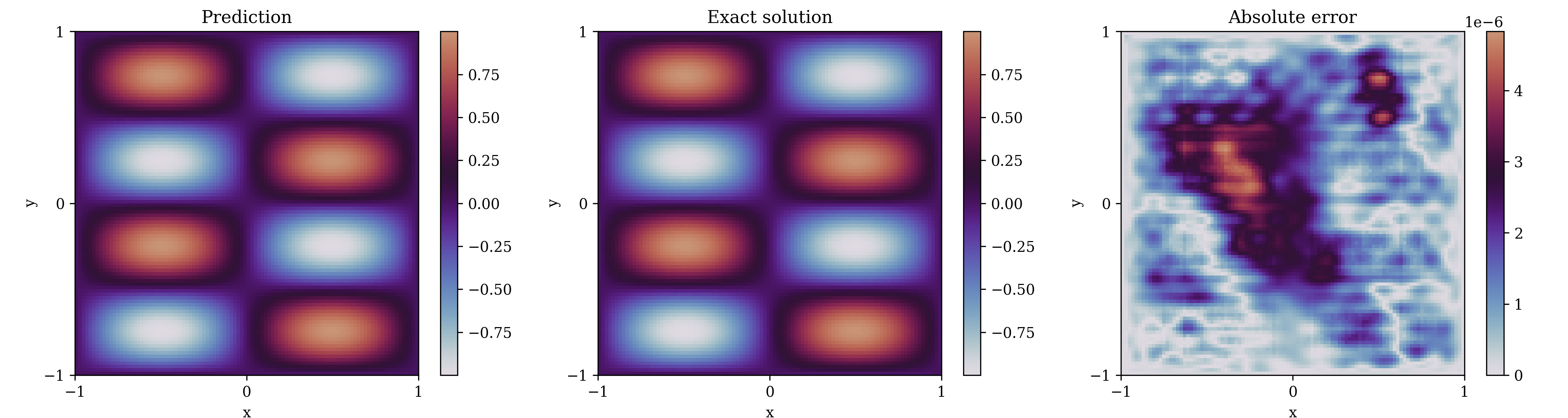}
    \caption{Helmholtz problem. \textbf{Left}: PI-Spline prediction. \textbf{Middle}: reference solution. \textbf{Right}: pointwise absolute error.}
    \label{fig:helmholtz_task}
\end{figure}

This choice produces a solution with a higher frequency content along the $y$-direction than along the $x$-direction. The PI-Spline approximation therefore uses a finer spline space than in the Poisson case, with splines of order $7$ and $25$ control points in each coordinate direction. The reference solution, PI-Spline prediction, and pointwise absolute error are shown in Figure~\ref{fig:helmholtz_task}.

\subsubsection{Exponential problem}
\label{subsubsec:exponential_problem}

We then turn to a manufactured elliptic problem on $\Omega=[0,1]\times[0,1]$, constructed to produce a solution with rapidly varying exponential features:
\begin{equation}
    \begin{aligned}
        \Delta u(x, y) + \frac{\partial u}{\partial x} = f(x, y) & \qquad \forall (x, y) \in \Omega, \\
        u(x, y) = 0 & \qquad \forall (x, y) \in \partial \Omega,
    \end{aligned}
\end{equation}
where the forcing term is defined as
\begin{equation}
    f(x, y) = (y -1)(e^{5y} -1)(6xe^{2x} - e^{2x} -1) + (x-1)(e^{2x} - 1)(25ye^{5y} - 15e^{5y}),
\end{equation}
yielding the analytical solution $u(x, y) = (x-1)(y-1)(e^{2x}-1)(e^{5y}-1)$.

\begin{figure}[ht]
    \centering
    \includegraphics[width=0.95\textwidth]{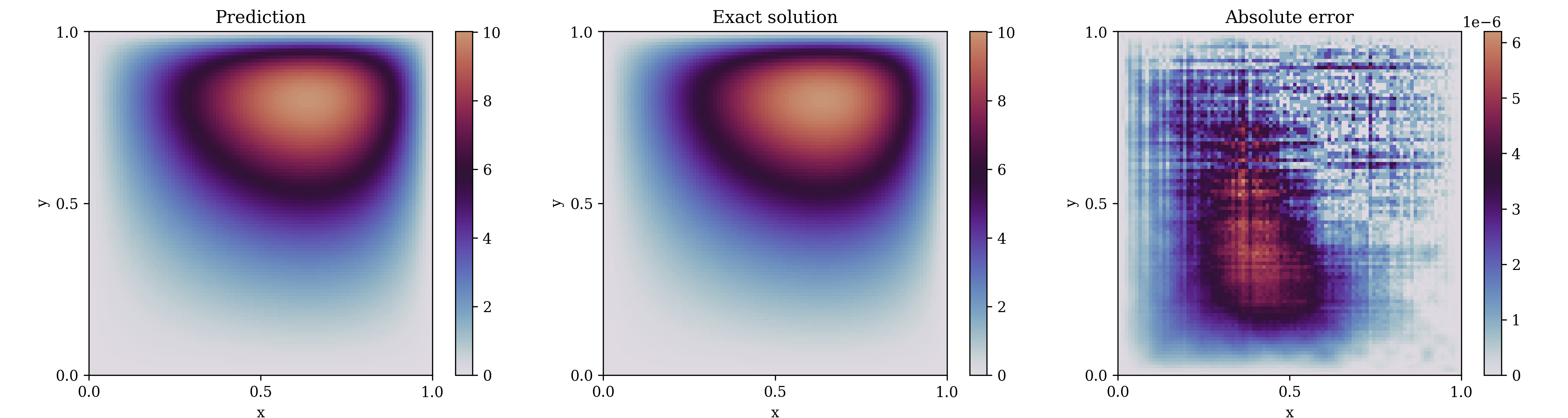}
    \caption{Exponential problem. \textbf{Left}: PI-Spline prediction. \textbf{Middle}: reference solution. \textbf{Right}: pointwise absolute error.}
    \label{fig:exponential_task}
\end{figure}

Unlike the previous benchmark, this problem is not oscillatory. Its difficulty instead comes from the rapidly varying exponential factors, which induce localized sharp gradients and a strongly inhomogeneous solution field. The PI-Spline approximation uses splines of order $7$ with $25$ control points in each coordinate direction. Figure~\ref{fig:exponential_task} reports the reference solution, the PI-Spline prediction, and the corresponding pointwise absolute error.

\subsubsection{Acoustic wave problem}
\label{subsubsec:wave_problem}

Finally, we consider a one-dimensional acoustic wave equation over the spatial domain $\Omega=[0,1]$ and the time interval $T=[0,1]$:
\begin{equation}
    \begin{aligned}
        \frac{\partial^2 u}{\partial t^2} - 4\frac{\partial^2 u}{\partial x^2} = 0 & \qquad \forall (x, t) \in \Omega \times T, \\
        u(x, 0) = \sin(\pi x) + \frac{1}{2}\sin(4 \pi x) & \qquad \forall x \in \Omega, \\
        \frac{\partial u}{\partial t}(x, 0) = 0 & \qquad \forall x \in \Omega, \\
        u(x, t) = 0 & \qquad \forall (x, t) \in \partial \Omega \times T.
    \end{aligned}
\end{equation}
The analytical solution reads $u(x, t) = \sin(\pi x)\cos(2\pi t) + \frac{1}{2}\sin(4\pi x)\cos(8\pi t)$.

\begin{figure}[ht]
    \centering
    \includegraphics[width=0.95\textwidth]{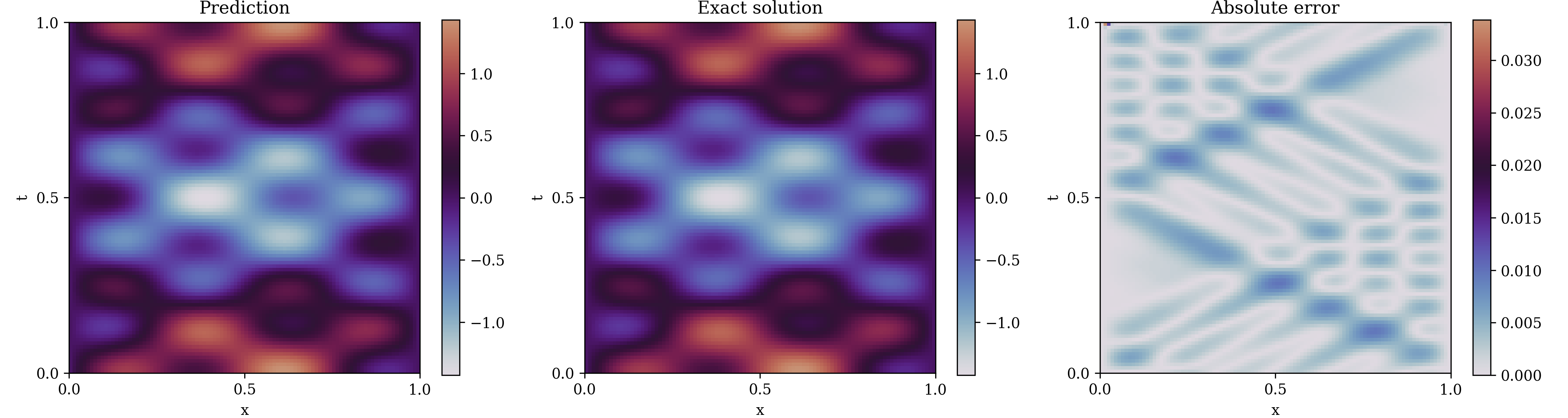}
    \caption{Acoustic wave problem. \textbf{Left}: PI-Spline prediction. \textbf{Middle}: reference solution. \textbf{Right}: pointwise absolute error.}
    \label{fig:wave_task}
\end{figure}

This benchmark is more challenging than the stationary problems because the approximation must resolve oscillatory behavior jointly in space and time. Moreover, as discussed in~\cite{WANG2022110768}, wave-equation PINNs may suffer from imbalanced convergence between the PDE residual and the initial and boundary constraints. We therefore use a weighted loss formulation for all methods, assigning weight $50$ to the initial displacement term, weight $10$ to the boundary and initial-velocity terms, and weight $1$ to the PDE residual. For PI-Splines, time is treated as an additional coordinate in the tensor-product representation. The model is trained for $10000$ epochs with extended LBFGS refinement, using splines of order $6$ with $40$ control points per coordinate direction. Spatial homogeneous Dirichlet boundary conditions are imposed strongly, while the initial conditions are retained as weak loss terms. The reference solution, PI-Spline prediction, and pointwise absolute error are shown in Figure~\ref{fig:wave_task}.

\subsection{Quantitative comparison}
\label{subsec:quantitative_results}

\begin{figure}[ht]
    \centering
    \includegraphics[width=0.9\textwidth]{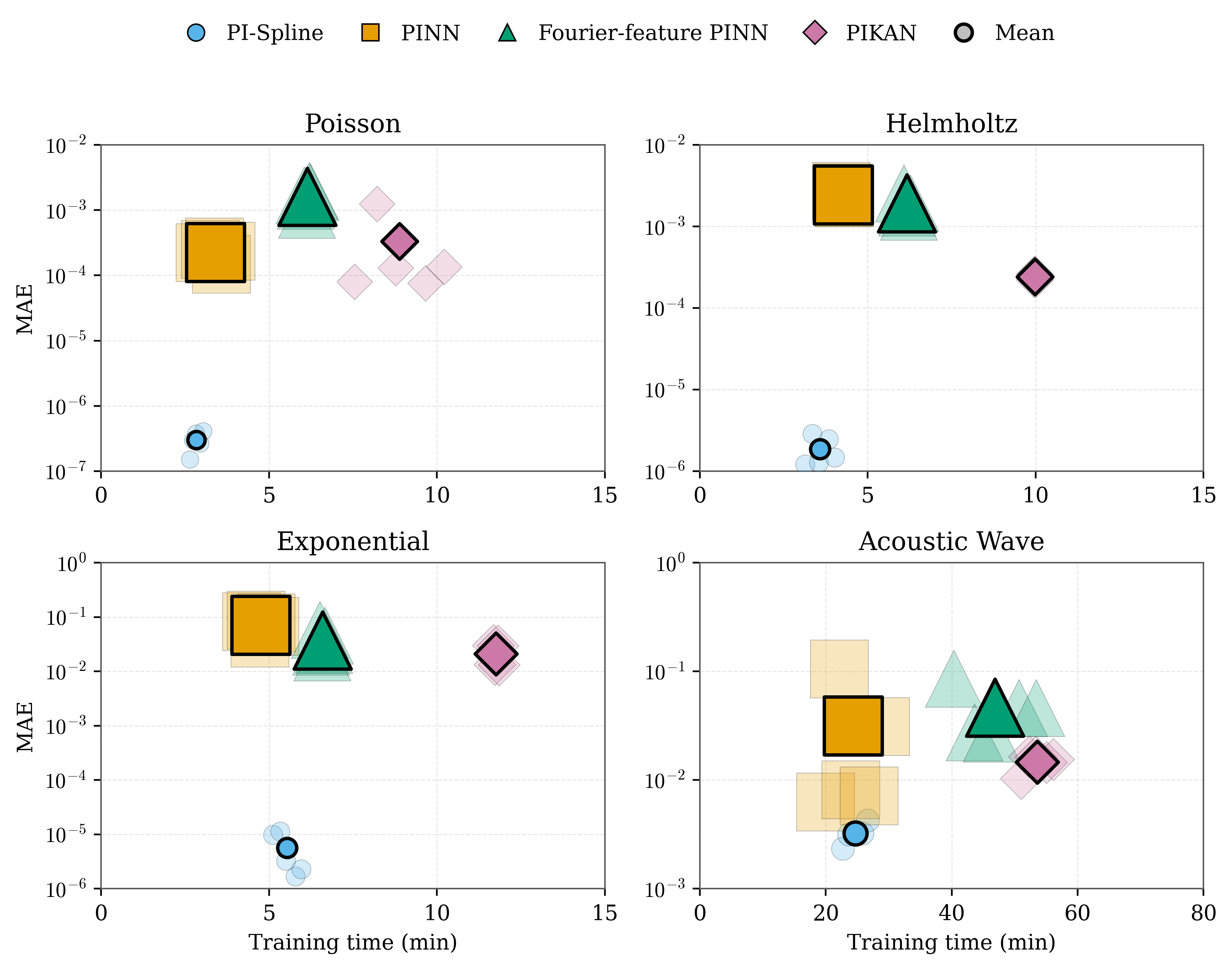}
    \caption{Accuracy--cost--complexity trade-off across benchmark problems. Each panel corresponds to one PDE benchmark. The horizontal axis reports training time, the vertical axis reports the mean absolute error, and marker size is proportional to the number of trainable parameters.}
    \label{fig:bubble_plot}
\end{figure}

A quantitative comparison across all benchmark problems is reported in Table~\ref{tab:results}. Results are given in terms of mean absolute error (MAE), number of trainable parameters, and training time, averaged over five independent runs. To complement the tabular comparison, Figure~\ref{fig:bubble_plot} provides a visual summary of the accuracy--cost--complexity trade-off.

\begin{table}
    \caption{Performance comparison across benchmark PDE tasks. Results are reported as mean $\pm$ standard deviation over 5 independent training runs. Best values for each task are in bold.}
    \vspace{4pt}
    \centering
    \begin{tabular}{llccc}
        \toprule
        Task & Model & MAE ($\downarrow$) & \# Parameters & Training Time (min) \\
        \midrule
        
        \multirow{4}{*}{Poisson} 
        & PI-Spline & $\mathbf{3.01^{\pm 1.01} \times 10^{-7}}$ & $\mathbf{169}$ & $\mathbf{2.84^{\pm 0.16}}$ \\
        & PINN & $2.26^{\pm 0.47} \times 10^{-4}$ & $17025$ & $3.41^{\pm 0.25}$ \\
        & PIKAN & $3.31^{\pm 5.06} \times 10^{-4}$ & $2049$ & $8.88^{\pm 1.07}$ \\
        & Fourier-feature PINN & $1.59^{\pm 0.39} \times 10^{-3}$ & 16641 & $6.14^{\pm 0.07}$\\
        
        \midrule
        
        \multirow{4}{*}{Helmholtz} 
        & PI-Spline & $\mathbf{1.86^{\pm 0.72} \times 10^{-6}}$ & $\mathbf{529}$ & $\mathbf{3.58^{\pm 0.36}}$ \\
        & PINN & $2.46^{\pm 0.19} \times 10^{-3}$ & $17025$ & $4.27^{\pm 0.04}$ \\
        & PIKAN & $2.42^{\pm 0.19} \times 10^{-4}$ & $2049$ & $9.98^{\pm 0.03}$ \\
        & Fourier-feature PINN & $1.92^{\pm 0.39} \times 10^{-3}$ & 16641 & $6.17^{\pm 0.08}$\\
        
        \midrule
        
        \multirow{4}{*}{Exponential} 
        & PI-Spline & $\mathbf{5.67^{\pm 4.40} \times 10^{-6}}$ & $\mathbf{529}$ & $5.54^{\pm 0.34}$ \\
        & PINN & $7.10^{\pm 1.84} \times 10^{-2}$ & $17025$ & $\mathbf{4.76^{\pm 0.22}}$ \\
        & PIKAN & $2.11^{\pm 0.82} \times 10^{-2}$ & $3393$ & $11.76^{\pm 0.07}$ \\
        & Fourier-feature PINN & $3.72^{\pm 1.43} \times 10^{-2}$ & 16641 & $6.59^{\pm 0.06}$\\
        
        \midrule
        
        \multirow{4}{*}{Wave} 
        & PI-Spline & $\mathbf{3.24^{\pm 0.68} \times 10^{-3}}$ & $\mathbf{1520}$ & $24.69^{\pm 1.58}$ \\
        & PINN & $3.15^{\pm 4.25} \times 10^{-2}$ & $17025$ & $\mathbf{24.35^{\pm 3.52}}$ \\
        & PIKAN & $1.45^{\pm 0.25} \times 10^{-2}$ & $3393$ & $53.58^{\pm 2.03}$ \\
        & Fourier-feature PINN & $4.64^{\pm 2.37} \times 10^{-2}$ & 16641 & $46.87^{\pm 5.26}$\\
        
        \bottomrule
    \end{tabular}
    \label{tab:results}
\end{table}

PI-Splines achieve the lowest error on all benchmark problems. The improvement is most pronounced for the stationary elliptic problems, where the spline representation provides substantially higher accuracy than the neural baselines while using markedly fewer trainable parameters. This behavior is evident in the Poisson and exponential cases, which respectively test smooth low-frequency approximation and localized sharp-gradient resolution. For the Helmholtz problem, PI-Splines retain a clear advantage despite the oscillatory structure of the solution.

The acoustic wave problem is the most challenging case. Although the relative gap is smaller than in the stationary benchmarks, PI-Splines still provide the most accurate approximation. This reduction in margin is expected, since the time-dependent problem requires the model to resolve oscillatory behavior in space and time while balancing residual, initial-condition, and boundary-condition terms.

In terms of model complexity, PI-Splines consistently use the smallest number of trainable parameters. This is reflected both in Table~\ref{tab:results} and in the marker sizes of Figure~\ref{fig:bubble_plot}. The result supports the main rationale of the proposed method: accuracy is obtained through a compact and structured approximation space rather than through neural overparametrization.

The training-time comparison shows that PI-Splines are also competitive in terms of runtime efficiency. Despite the overhead associated with evaluating tensor-product B-spline bases and their derivatives, PI-Splines are the fastest method in two of the four benchmarks and remain close to standard PINNs in the remaining cases. They are also consistently faster than PIKANs and Fourier-feature PINNs, while achieving the lowest approximation errors across all benchmarks.

Overall, the results indicate that PI-Splines provide a favorable accuracy--complexity trade-off across the considered benchmarks. The method combines low parameter counts with consistently improved accuracy, while retaining training times that are competitive with the neural alternatives once the cost of spline-basis evaluation is taken into account.

\subsection{Sensitivity analysis}
\label{subsec:sensitivity_analysis}

\begin{figure}[ht]
\centering
    \includegraphics[width=0.9\textwidth]{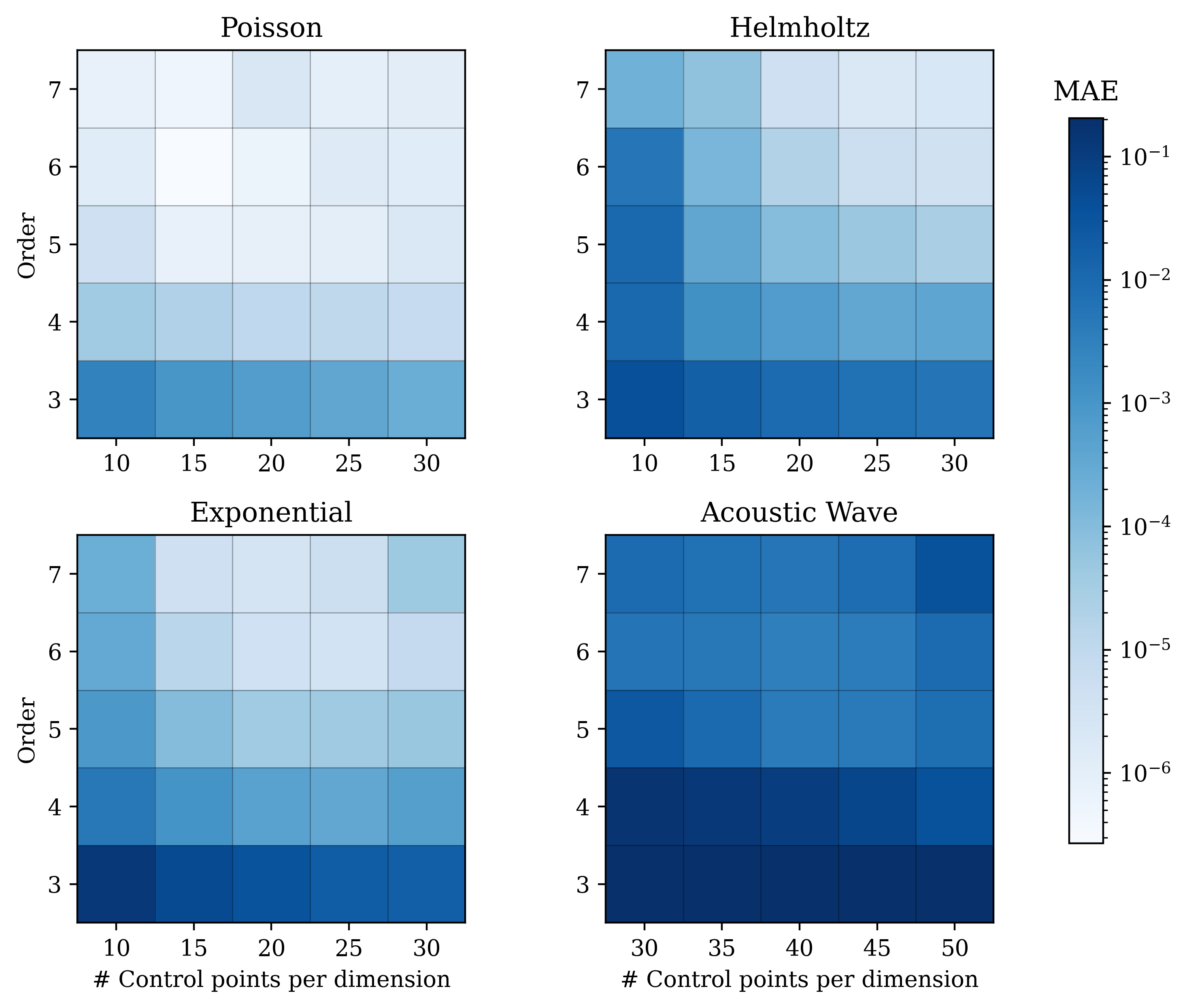}
    \caption{Mean absolute error (MAE) across spline configurations for all benchmark problems. Each heatmap reports the MAE as a function of spline order and number of control points per dimension, illustrating the impact of spline resolution and smoothness on approximation accuracy across problems of increasing complexity.}
    \label{fig:heatmaps}
\end{figure}

To assess the dependence of PI-Splines on their main discretization parameters, we perform a grid search over spline order and number of control points per coordinate direction. Figure~\ref{fig:heatmaps} reports the resulting MAE for each benchmark problem. These two parameters control distinct aspects of the approximation space: the number of control points determines the spatial or space-time resolution of the spline space, whereas the spline order determines the local polynomial degree and the maximum continuity across knot spans.

The heatmaps show broader low-error regions for the Poisson problem, whereas the Helmholtz, exponential, and wave problems require higher control-point counts to reach comparable accuracy. Across all tasks, increasing the number of control points is the dominant factor in reducing the error, while increasing the spline order produces more problem-dependent gains. Higher orders can improve smoothness and local approximation power, but the benefits diminish once sufficient resolution is available.

Overall, the analysis indicates that PI-Spline performance is primarily resolution-limited for the problems considered here. The number of control points should therefore be selected according to the smallest relevant spatial or temporal scale in the solution, while the spline order should provide sufficient smoothness for the differential operator without unnecessarily increasing computational cost.

\section{Conclusions}\label{sec:conclusions}
In this work, we introduced PI-Splines, a structured spline-based alternative to classical neural architectures for physics-informed learning. The method retains the residual-based optimization framework of PINNs, but replaces the neural parametrization with a tensor-product B-spline expansion whose control coefficients are optimized directly. This construction combines physics-informed training with compact support, explicit smoothness control, analytical differentiation, and a direct geometric interpretation of the trainable parameters.

The numerical results show that PI-Splines provide a favorable accuracy--complexity trade-off on the considered benchmarks. Across the Poisson, Helmholtz, exponential, and acoustic wave equations, the method consistently achieves the lowest approximation error among the tested physics-informed models. The improvement is especially pronounced in the stationary elliptic problems, including the case with localized sharp gradients, where the locality of the B-spline basis supports the representation of spatially confined solution features. At the same time, PI-Splines require substantially fewer trainable parameters than standard PINNs, Fourier-feature PINNs, and PIKANs, with reductions of up to two orders of magnitude relative to the tested baselines.

This reduction in parameter count, however, does not necessarily translate into proportional reductions in training time. Each optimization step requires evaluating tensor-product B-spline basis functions and their derivatives at the collocation points, which introduces a non-negligible computational cost, particularly for dense collocation sets, high spline orders, or higher-dimensional tensor-product spaces. Thus, the efficiency of PI-Splines depends not only on the number of optimized coefficients, but also on the implementation of basis evaluation and derivative computation.

The main limitation of the method is its dependence on the prescribed spline discretization. For a fixed knot vector, spline order, and number of control points, the approximation space is finite-dimensional and may be too restrictive if the resolution is insufficient. Conversely, increasing the number of control points enlarges the approximation space but also increases computational cost. The sensitivity analysis confirms that, for the problems considered here, the number of control points is the dominant factor controlling accuracy, while the effect of spline order is more problem-dependent.

These observations suggest that PI-Splines are best viewed as a complementary physics-informed approximation framework rather than a universal replacement for PINNs or classical discretization methods. They are particularly suitable for moderate-dimensional problems with sufficiently regular or piecewise-smooth solutions, where localized structures must be represented accurately without relying on heavily overparametrized neural architectures.

Future work will focus on improving efficiency, adaptivity, and geometric flexibility. Promising directions include sparse and hardware-aware basis evaluation, precomputed basis matrices, adaptive knot or control-point refinement, and anisotropic refinement for direction-dependent features. Extensions to complex geometries could exploit NURBS-based parametrizations and connections with isogeometric analysis, while time-dependent problems may benefit from adaptive space--time discretizations or autoregressive formulations over control coefficients.

Overall, PI-Splines offer a promising route for physics-informed learning beyond standard neural-network architectures. By combining residual-based training with a structured spline approximation space, the method provides an interpretable and parameter-efficient framework that achieves high accuracy on representative differential problems while preserving clear avenues for further algorithmic development.

\section{Acknowledgements}
G.C. acknowledges the support provided by the European Union -- NextGenerationEU under the National Recovery and Resilience Plan (PNRR), Mission 4, Component 2, Investment 3.3, through a PhD scholarship at Scuola Internazionale Superiore di Studi Avanzati di Trieste (CUP G93C24000530001), carried out in collaboration with Fincantieri S.p.A.
N.D. and G.R. acknowledge the support provided by the European Union – NextGenerationEU within the framework of the iNEST – Interconnected Nord-Est Innovation Ecosystem project (iNEST ECS00000043; CUP G93C22000610007) and its CC5 Young Researchers initiative.
The views and opinions expressed are solely those of the authors and do not necessarily reflect those of the European Union. The European Union cannot be held responsible for them.

\bibliographystyle{abbrv}
\bibliography{bibliography}

@article{coscia2023physics,
    title={Physics-Informed Neural networks for Advanced modeling},
    author={Coscia, Dario and Ivagnes, Anna and Demo, Nicola and Rozza, Gianluigi},
    journal={Journal of Open Source Software},
    volume={8},
    number={87},
    pages={5352},
    year={2023}
}

@article{RAISSI2019686,
    title={Physics-informed neural networks: A deep learning framework for solving forward and inverse problems involving nonlinear partial differential equations},
    journal={Journal of Computational Physics},
    volume={378},
    pages={686-707},
    year={2019},
    issn={0021-9991},
    doi={https://doi.org/10.1016/j.jcp.2018.10.045},
    url={https://www.sciencedirect.com/science/article/pii/S0021999118307125},
    author = {M. Raissi and P. Perdikaris and G.E. Karniadakis},
}

@inproceedings{liu2025kan,
    title={{KAN}: Kolmogorov{\textendash}Arnold Networks},
    author={Ziming Liu and Yixuan Wang and Sachin Vaidya and Fabian Ruehle and James Halverson and Marin Soljacic and Thomas Y. Hou and Max Tegmark},
    year={2025},
    url={https://proceedings.iclr.cc/paper_files/paper/2025/file/afaed89642ea100935e39d39a4da602c-Paper-Conference.pdf},
    volume={2025},
    booktitle={International Conference on Learning Representations},
    editor={Y. Yue and A. Garg and N. Peng and F. Sha and R. Yu},
    pages={70367--70413},
}

@article{Toscano2024FromPT,
    title={From {PINN}s to {PIKAN}s: recent advances in physics-informed machine learning},
    author={Juan Diego Toscano and Vivek Oommen and Alan John Varghese and Zongren Zou and Nazanin Ahmadi Daryakenari and Chenxi Wu and George Em. Karniadakis},
    journal={Machine Learning for Computational Science and Engineering},
    year={2024},
    volume={1},
    url={https://api.semanticscholar.org/CorpusID:273403632}
}

@inproceedings{2015-kingma,
    title={Adam: A Method for Stochastic Optimization.},
    author={Diederik P. Kingma and Jimmy Ba},
    booktitle={3rd International Conference on Learning Representations},
    url={http://dblp.uni-trier.de/db/conf/iclr/iclr2015.html#KingmaB14},
    year={2015},
    pages={},
}

@article{SI2026114713,
    title={Complex physics-informed neural network},
    journal={Journal of Computational Physics},
    volume={553},
    pages={114713},
    year={2026},
    issn={0021-9991},
    doi={https://doi.org/10.1016/j.jcp.2026.114713},
    url={https://www.sciencedirect.com/science/article/pii/S002199912600063X},
    author = {Chenhao Si and Ming Yan and Xin Li and Zhihong Xia},
}

@article{WANG2022110768,
    title={When and why {PINN}s fail to train: A neural tangent kernel perspective},
    journal={Journal of Computational Physics},
    volume={449},
    pages={110768},
    year={2022},
    issn={0021-9991},
    doi={https://doi.org/10.1016/j.jcp.2021.110768},
    url={https://www.sciencedirect.com/science/article/pii/S002199912100663X},
    author = {Sifan Wang and Xinling Yu and Paris Perdikaris},
}

@article{doi:10.1137/20M1318043,
    author={Sifan Wang and Yujun Teng and Paris Perdikaris},
    title={Understanding and Mitigating Gradient Flow Pathologies in Physics-Informed Neural Networks},
    journal={SIAM Journal on Scientific Computing},
    volume={43},
    number={5},
    pages={A3055-A3081},
    year={2021},
    doi={10.1137/20M1318043},
    URL={https://doi.org/10.1137/20M1318043}
}

@inproceedings{Rahaman2018OnTS,
  title={On the Spectral Bias of Neural Networks},
  author={Nasim Rahaman and Aristide Baratin and Devansh Arpit and Felix Dr{\"a}xler and Min Lin and Fred A. Hamprecht and Yoshua Bengio and Aaron C. Courville},
  pages={},
  booktitle={International Conference on Machine Learning},
  year={2019},
  url={https://api.semanticscholar.org/CorpusID:53012119}
}

@article{MCCLENNY2023111722,
    title={Self-adaptive physics-informed neural networks},
    journal={Journal of Computational Physics},
    volume={474},
    pages={111722},
    year={2023},
    issn={0021-9991},
    doi={https://doi.org/10.1016/j.jcp.2022.111722},
    url={https://www.sciencedirect.com/science/article/pii/S0021999122007859},
    author = {Levi D. McClenny and Ulisses M. Braga-Neto},
}

@article{ANAGNOSTOPOULOS2024116805,
    title={Residual-based attention in physics-informed neural networks},
    journal={Computer Methods in Applied Mechanics and Engineering},
    volume={421},
    pages={116805},
    year={2024},
    issn={0045-7825},
    doi={https://doi.org/10.1016/j.cma.2024.116805},
    url={https://www.sciencedirect.com/science/article/pii/S0045782524000616},
    author={Sokratis J. Anagnostopoulos and Juan Diego Toscano and Nikolaos Stergiopulos and George Em Karniadakis},
}

@book{splinebook,
    author={Carl de Boor},
    year={1978},
    month={01},
    pages={},
    title={A Practical Guide to Splines},
    volume={Volume 27},
    journal={Applied Mathematical Sciences, New York: Springer, 1978},
    publisher={Springer},
    doi={10.2307/2006241}
}

@book{nurbsbook,
    author={Les Piegl and Wayne Tiller},
    year={1996},
    month={11},
    pages={646},
    title={The {NURBS} Book},
    volume={},
    publisher={Springer},
    doi={https://doi.org/10.1007/978-3-642-59223-2}
}

@article{karniadakisphysics2021,
    author={George Em Karniadakis and Ioannis G. Kevrekidis and Lu Lu and Paris Perdikaris and Sifan Wang and Liu Yang},
    title={Physics-informed machine learning},
    journal={Nature Reviews Physics},
    year={2021},
    volume={3},
    issn={2522-5820},
    doi={10.1038/s42254-021-00314-5},
    url={https://doi.org/10.1038/s42254-021-00314-5},
}

@article{pinns_vs_FEM,
    author={Grossmann, Tamara and Komorowska, Urszula and Latz, Jonas and Schönlieb, Carola-Bibane},
    year={2024},
    month={05},
    pages={},
    title={Can physics-informed neural networks beat the finite element method?},
    volume={89},
    journal={IMA Journal of Applied Mathematics},
    doi={10.1093/imamat/hxae011}
}

@article{daw2023mitigating,
    title={Mitigating Propagation Failures in Physics-informed Neural Networks using Retain-Resample-Release ({R}3) Sampling}, 
    author={Arka Daw and Jie Bu and Sifan Wang and Paris Perdikaris and Anuj Karpatne},
    year={2023},
    journal={ArXiv},
    volume={abs/2207.02338}
}

@article{jagtap2020extended,
    title={Extended physics-informed neural networks (xpinns): A generalized space-time domain decomposition based deep learning framework for nonlinear partial differential equations},
    author={Ameya D. Jagtap and George Em Karniadakis},
    journal={Communications in Computational Physics},
    volume={28},
    number={5},
    pages={2002--2041},
    year={2020}
}

@article{Wang2024PirateNetsPD,
  title={{PirateNets}: Physics-informed Deep Learning with Residual Adaptive Networks},
  author={Sifan Wang and Bowen Li and Yuhan Chen and Paris Perdikaris},
  journal={ArXiv},
  year={2024},
  volume={abs/2402.00326},
}

@article{Wang2023AnEG,
  title={An Expert's Guide to Training Physics-informed Neural Networks},
  author={Sifan Wang and Shyam Sankaran and Hanwen Wang and Paris Perdikaris},
  journal={ArXiv},
  year={2023},
  volume={abs/2308.08468},
}

@article{WANG2021113938,
    title={On the eigenvector bias of Fourier feature networks: From regression to solving multi-scale {PDE}s with physics-informed neural networks},
    journal={Computer Methods in Applied Mechanics and Engineering},
    volume={384},
    pages={113938},
    year={2021},
    issn={0045-7825},
    doi={https://doi.org/10.1016/j.cma.2021.113938},
    author={Sifan Wang and Hanwen Wang and Paris Perdikaris},
}

@article{
    wang2026physicsinformed,
    title={Physics-Informed Deep B-Spline Networks},
    author={Zhuoyuan Wang and Raffaele Romagnoli and Saviz Mowlavi and Yorie Nakahira},
    journal={Transactions on Machine Learning Research},
    issn={2835-8856},
    year={2026},
    url={https://openreview.net/forum?id=tHO2zEqmzm},
    note={}
}

@article{
    hermite,
    author={Wandel, Nils and Weinmann, Michael and Neidlin, Michael and Klein, Reinhard},
    year={2022},
    month={06},
    pages={8529-8538},
    title={Spline-{PINN}: Approaching PDEs without Data Using Fast, Physics-Informed Hermite-Spline {CNN}s},
    volume={36},
    journal={Proceedings of the AAAI Conference on Artificial Intelligence},
    doi={10.1609/aaai.v36i8.20830}
}

@article{
    SUN2023110165,
    title={{PiSL}: Physics-informed Spline Learning for data-driven identification of nonlinear dynamical systems},
    journal={Mechanical Systems and Signal Processing},
    volume={191},
    pages={110165},
    year={2023},
    issn={0888-3270},
    doi={https://doi.org/10.1016/j.ymssp.2023.110165},
    url={https://www.sciencedirect.com/science/article/pii/S0888327023000729},
    author={Fangzheng Sun and Yang Liu and Qi Wang and Hao Sun},
}

@article{
    deepnurbs,
    title={Deep {NURBS}—admissible physics-informed neural networks},
    journal={Engineering with Computers},
    volume={40},
    pages={4007-4021},
    year={2024},
    doi={https://doi.org/10.1007/s00366-024-02040-9},
    author={Hamed Saidaoui and Luis Espath and Raul Tempone}
}

@article{
    tamburlin2026constraint,
    title = {Constraint-driven Optimization and Parametrization of Industrial {NURBS} Geometries via Neural Deformation Field},
    author = {Federico Tamburlin and Giovanni Canali and Giuseppe Alessio D'Inverno and Nicola Demo and Andrea Mola and Gianluigi Rozza},
    journal={ArXiv},
    year={2026},
    volume={abs/2606.07198},
    url={https://arxiv.org/abs/2606.07198}, 
}

\appendix
\setcounter{secnumdepth}{1}
\section{Training details}
\label{sec:app_A}

All models were trained using a common physics-informed pipeline, with the aim of isolating the effect of the approximation architecture. For each benchmark problem, the governing equation, collocation sets, loss terms, optimization schedule, and random seeds were kept fixed across the compared methods whenever applicable. Each experiment was repeated over five independent random seeds, and the results reported in the main text are given as mean and standard deviation over these runs. The common training settings are summarized in Table~\ref{tab:training_common}.

\begin{table}[H]
\centering
\caption{Common training settings used across all architectures.}
\label{tab:training_common}
    \begin{tabular}{ll}
    \toprule
    Setting & Value \\
    \midrule
    Random seeds & $1,2,3,5,8$ \\
    Optimizer & Adam $+$ LBFGS \\
    Adam learning rate & $10^{-3}$ \\
    Learning-rate scheduler & MultiStepLR \\
    Scheduler milestone & $2000$ epochs \\
    Scheduler decay factor & $0.1$ \\
    LBFGS line search & Strong Wolfe \\
    Interior collocation points & $5000$ \\
    Boundary collocation points & $500$ \\
    Initial-time points (wave problem) & $1000$ \\
    \bottomrule
    \end{tabular}
\end{table}

The optimization schedule was chosen according to the benchmark class. Stationary problems were trained for $5000$ epochs, whereas the acoustic wave problem required a longer training horizon and an extended LBFGS refinement phase. The corresponding schedules are reported in Table~\ref{tab:training_schedule}.

\begin{table}[H]
\centering
\caption{Problem-dependent optimization schedule.}
\label{tab:training_schedule}
    \begin{tabular}{lccc}
    \toprule
    Problems & Total epochs & Adam epochs & LBFGS epochs \\
    \midrule
    Poisson, Helmholtz, Exponential & $5000$ & $4900$ & $100$ \\
    Acoustic wave & $10000$ & $8500$ & $1500$ \\
    \bottomrule
    \end{tabular}
\end{table}

The stationary problems were trained using an unweighted residual formulation. For the acoustic wave equation, loss weights were introduced to balance the PDE residual with the boundary and initial-condition terms. The PDE residual was assigned weight $1$, the initial displacement term weight $50$, and both the initial-velocity and boundary-condition terms weight $10$. The same weighting strategy was used for all architectures in the wave benchmark.

The standard PINN baseline was implemented as a fully connected feed-forward neural network with two hidden layers of width $128$. The Fourier-feature PINN used the same feed-forward architecture, preceded by a single Fourier-feature embedding layer of width $128$. For the stationary problems, the Fourier scales were set to $\sigma=\{0.25,0.5,1.0,2.0\}$, while for the acoustic wave problem they were set to $\sigma=\{0.5,1.0,2.0,4.0\}$.

The physics-informed KAN baseline used a single hidden layer of width $32$, since this configuration led to more stable training, with spline order $6$ and grid range $[-2,2]$. The number of knots was set to $20$ for the Poisson and Helmholtz problems, and to $40$ for the exponential and acoustic wave problems. During KAN training, an adaptive grid update was performed every $100$ epochs before the LBFGS phase.

For PI-Splines, the solution was represented by a tensor-product B-spline expansion. The number of knots in each coordinate direction was computed as the number of control points plus the spline order. The configurations used in the main quantitative comparison were selected from the grid search performed in the sensitivity analysis and are reported in Table~\ref{tab:spline_main_configs}.

\begin{table}[H]
\centering
\caption{PI-Spline configurations used for the main quantitative comparison.}
\label{tab:spline_main_configs}
    \begin{tabular}{lcc}
    \toprule
    Problem & Spline order & Control points per direction \\
    \midrule
    Poisson & $6$ & $15$ \\
    Helmholtz & $7$ & $25$ \\
    Exponential & $7$ & $25$ \\
    Acoustic wave & $6$ & $40$ \\
    \bottomrule
    \end{tabular}
\end{table}

After training, the mean absolute error was computed by comparing the predicted solution with the analytical solution on independently sampled evaluation points. For each run, the physical residual was evaluated on the same set of points. The reported training times include both the Adam and LBFGS phases.

\end{document}